\documentclass[11pt,letterpaper]{article}
\usepackage[hyperref]{emnlp2017}
\usepackage{times}
\usepackage{latexsym}

\usepackage[T2A, T1]{fontenc}
\usepackage[utf8]{inputenc}

\usepackage{url}
\usepackage{graphicx}
\usepackage{multirow}
\usepackage{amsthm}
\usepackage{amssymb}
\usepackage{amsmath}
\usepackage{amsfonts}
\usepackage{color}
\usepackage{mdwlist}
\usepackage{microtype}
\usepackage{tabularx}
\usepackage{booktabs}
\usepackage{subcaption}
\usepackage{color,colortbl}

\definecolor{Gray}{gray}{0.9}

\newcolumntype{b}{X}
\newcolumntype{m}{>{\hsize=.6\hsize}X}
\newcolumntype{s}{>{\hsize=.33\hsize}X}

\makeatletter
\newcommand{\removelatexerror}{\let\@latex@error\@gobble}

\makeatother

\usepackage[russian,english]{babel}

\emnlpfinalcopy % Uncomment this line for the final submission
\title{Cross-Lingual Induction and Transfer of Verb Classes \\ Based on Word Vector Space Specialisation}

\author{Ivan Vuli\'c$^{\mathbf{1}}$ ,~ Nikola Mrk\v{s}i\'{c}$^{\mathbf{2}}$ ~{\normalfont and} {Anna Korhonen}$^{\mathbf{1}}$  \\
$^{\mathbf{1}}$ Language Technology Lab, University of Cambridge, UK  \\
$^{\mathbf{2}}$ Dialogue Systems Group, University of Cambridge, UK \\
\texttt{\{iv250,nm480,alk23\}@cam.ac.uk}
}

\date{}

\begin{document}

\maketitle

\begin{abstract}
Existing approaches to automatic VerbNet-style verb classification are heavily dependent on feature engineering and therefore limited to languages with mature NLP pipelines. In this work, we propose a novel cross-lingual transfer method for inducing VerbNets for multiple languages. To the best of our knowledge, this is the first study which demonstrates how the architectures for learning word embeddings can be applied to this challenging syntactic-semantic task. Our method uses cross-lingual translation pairs to tie each of the six target languages into a bilingual vector space with English, jointly specialising the representations to encode the relational information from English VerbNet. A standard clustering algorithm is then run on top of the VerbNet-specialised representations, using vector dimensions as features for learning verb classes. Our results show that the proposed cross-lingual transfer approach sets new state-of-the-art verb classification performance across all six target languages explored in this work. %, yielding considerable improvements.
\end{abstract}

\section{Introduction}
\label{s:intro}
% About verbs
Playing a key role in conveying the meaning of a sentence, verbs are famously complex. They  
display a wide range of syntactic-semantic behaviour, expressing the semantics of an event as well as relational information among its participants \cite[inter alia]{Jackendoff:1972book,Gruber:1976book,Levin:1993book}.

%Jackendoff:1992book

% About importance of lexical resources and VerbNet
Lexical resources which capture the variability of verbs are instrumental for many Natural Language Processing (NLP) applications. One of the richest verb resources currently available for English is VerbNet \cite{Kipper:2000aaai,Kipper:2005thesis}.\footnote{http://verbs.colorado.edu/$\sim$mpalmer/projects/verbnet.html} Based on the work of Levin \shortcite{Levin:1993book}, this largely hand-crafted taxonomy organises verbs into classes on the basis of their shared syntactic-semantic behaviour. Providing a useful level of generalisation for many NLP tasks, VerbNet has been used to support semantic role labelling \cite{Swier:2004emnlp,Giuglea:2006acl}, semantic parsing \cite{Shi:2005cicling}, word sense disambiguation \cite{Brown:2011iwcs}, discourse parsing \cite{Subba:2009naacl}, information extraction \cite{Mausam:2012emnlp}, text mining applications \cite{Lippincott:2013jbi,Rimell:2013jbi}, research into human language acquisition \cite{Korhonen:2010}, and other tasks.

% Its cross-linguistic component
This benefit for English NLP has motivated the development of VerbNets for languages such as Spanish and Catalan \cite{Aparicio:2008lrec}, Czech \cite{Pala:2008raslan}, and Mandarin \cite{Liu:2008ll}. However,  end-to-end manual resource development using Levin's methodology is extremely time consuming, even when supported by translations of English VerbNet classes to other languages \cite{Sun:2010coling,Scarton:2014cicling}. Approaches which aim to learn verb classes automatically offer an attractive alternative. However, existing methods rely on carefully engineered features that are extracted using sophisticated language-specific resources \cite[i.a.]{Joanis:2008nle,Sun:2010coling,Falk:2012acl}, ranging from accurate parsers to pre-compiled subcategorisation frames \cite{Schulte:2006cl,Li:2008acl,Messiant:2008acl}. Such methods are limited to a small set of resource-rich languages.  
 
It has been argued that VerbNet-style classification has a strong cross-lingual element \cite{Jackendoff:1992book,Levin:1993book}. In support of this argument, \newcite{Majewska:2017lre} have shown that English VerbNet has high translatability across different, even typologically diverse languages. Based on this finding, we propose an automatic approach which exploits readily available annotations for English to facilitate efficient, large-scale development of VerbNets for a wide set of {target} languages.

Recently, unsupervised methods for inducing distributed word vector space representations or \textit{word embeddings} \cite{Mikolov:2013iclr} have been successfully applied to a plethora of NLP tasks \cite[i.a.]{Turian:2010acl,Collobert:2011jmlr,Baroni:2014acl}. These methods offer an elegant way to learn directly from large corpora, bypassing the feature engineering step and the dependence on mature NLP pipelines (e.g., POS taggers, parsers, extraction of subcategorisation frames). In this work, we demonstrate how these models can be used to support automatic verb class induction. Moreover, we show that these models offer the means to exploit  inherent cross-lingual links in VerbNet-style classification in order to guide the development of new classifications for resource-lean languages. To the best of our knowledge, this proposition has not been investigated in previous work.

There has been little work on assessing the suitability of embeddings for capturing rich syntactic-semantic phenomena. One challenge is their reliance on the {distributional hypothesis} \cite{Harris:1954}, which coalesces fine-grained syntactic-semantic relations between words into a broad relation of semantic relatedness (e.g., \textit{coffee:cup}) \cite{Hill:2015cl,Kiela:2015emnlp}. This property has an adverse effect when word embeddings are used in downstream tasks such as spoken language understanding \cite{Kim:16b,Kim:16} or dialogue state tracking \cite{Mrksic:16,Mrksic:16b}. It could have a similar effect on verb classification, which relies on the similarity in syntactic-semantic properties of verbs within a class. In summary, we explore three important questions in this paper:

{\bf (Q1)} Given their fundamental dependence on the distributional hypothesis, to what extent can unsupervised methods for inducing vector spaces facilitate the automatic induction of VerbNet-style verb classes across different languages?

{\bf (Q2)} Can one boost verb classification for lower-resource languages by exploiting general-purpose cross-lingual resources such as BabelNet \cite{Navigli:12,Ehrmann:14} or bilingual dictionaries such as PanLex \cite{Kamholz:2014lrec} to construct better word vector spaces for these languages?

{\bf (Q3)} Based on the stipulated cross-linguistic validity of VerbNet-style classification, can one exploit rich sets of readily available annotations in one language (e.g., the full English VerbNet) to automatically bootstrap the creation of VerbNets for other languages? In other words, is it possible to exploit a cross-lingual vector space to transfer VerbNet knowledge from a resource-rich to a resource-lean language?

To investigate Q1, we induce standard distributional vector spaces \cite{Mikolov:2013nips,Levy:2014acl} from large monolingual corpora in English and six target languages. As expected, the results obtained with this straightforward approach show positive trends, but at the same time reveal its limitations for all the languages involved. Therefore, the focus of our work shifts to Q2 and Q3. The problem of inducing VerbNet-oriented embeddings is framed as vector space specialisation using the available external resources: BabelNet or PanLex, and (English) VerbNet. Formalised as an instance of post-processing \textit{semantic specialisation} approaches \cite{Faruqui:2015naacl,Mrksic:16}, our procedure is steered by two sets of linguistic constraints: \textbf{1)} cross-lingual (translation) links between languages extracted from BabelNet (targeting Q2); and \textbf{2)} the available VerbNet annotations for a resource-rich language. The two sets of constraints jointly target Q3.

\begin{figure*}[t]
\centering
\includegraphics[width=0.77\linewidth]{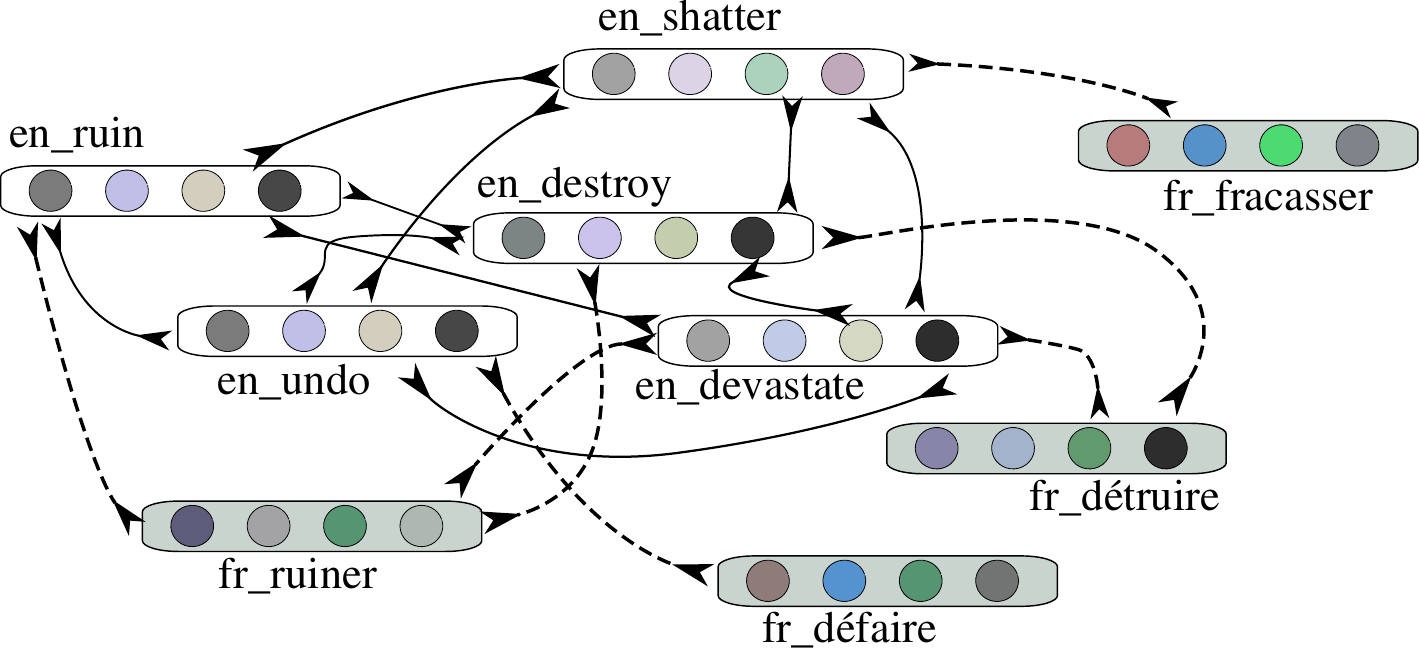}
\vspace{-0.4em}
\caption{Transferring VerbNet information from a resource-rich to a resource-lean language through a word vector space: an \textit{English}$ \rightarrow$ \textit{French} toy example. Representations of words described by two types of \textsc{Attract} constraints are pulled closer together in the joint vector space. (1) Monolingual pairwise constraints in English (e.g., \textit{(en\_ruin, en\_shatter), (en\_destroy, en\_undo)}) reflect the \textsc{en} VerbNet structure and are generated from the readily available verb classification in English (solid lines). They are used to specialise the distributional \textsc{en} vector subspace for the VerbNet relation. (2) Cross-lingual English-French pairwise constraints (extracted from BabelNet) describe cross-lingual synonyms (i.e., translation links) such as \textit{(en\_ruin, fr\_ruiner) or (en\_shatter, fr\_fracasser)} (dashed lines). The post-processing fine-tuning specialisation procedure based on (1) and (2) effectively transforms the initial distributional French vector subspace to also emphasise the VerbNet-style structure, facilitating the induction of verb classes in French.}
\vspace{-1.1mm}
\label{fig:illustration}
\end{figure*}

The main goal of vector space specialisation is to pull examples standing in desirable relations, as described by the constraints, closer together in the transformed vector space. The specialisation process can capitalise on the knowledge of VerbNet relations in the \emph{source} language (English) by using translation pairs to transfer that knowledge to each of the \emph{target} languages. By constructing  shared bilingual vector spaces, our method facilitates the transfer of semantic relations derived from VerbNet to the vector spaces of resource-lean target languages. This idea is illustrated by Fig.~\ref{fig:illustration}.

Our results indicate that cross-lingual connections yield improved verb classes across all six target languages (thus answering Q2). Moreover, a consistent and significant boost in verb classification performance is achieved by propagating the VerbNet-style information from the source language (English) to any other target language (e.g., Italian, Croatian, Polish, Finnish) for which no VerbNet-style information is available during the fine-tuning process (thus answering Q3).\footnote{On a high level, we demonstrate that a constraints-driven fine-tuning framework can specialise word embeddings to reflect VerbNet-style relations which rely not only on verb sense similarity, but also on similarity in syntax, selectional preferences, and diathesis alternations.} We report state-of-the-art verb classification performance for all six languages in our experiments. For instance, we improve the state-of-the-art F-1 score from prior work from 0.55 to 0.79 for French, and from 0.43 to 0.74 for Brazilian Portuguese.

\section{Methodology: Specialising for VerbNet}
\label{s:methodology}
\paragraph{Motivation: Verb Classes and VerbNet}
VerbNet is a hierarchical, domain-independent, broad-coverage verb lexicon based on Levin's classification and taxonomy of English (\textsc{en}) verbs \cite{Levin:1993book,Kipper:2005thesis}. Verbs are grouped into classes (e.g. the class \textsc{put-9.1} for verbs such as \textit{place}, \textit{position}, \textit{insert}, and \textit{arrange}) based on their shared meaning components and syntactic behaviour, defined in terms of their participation in \textit{diathesis alternations}, i.e., alternating verb frames that are related with the same or similar meaning. VerbNet extends and refines Levin's classification, providing more fine-grained syntactic and semantic information for individual classes.  Each VerbNet class is characterised by its member verbs, syntactic frames, semantic predicates and typical verb arguments.\footnote{The usefulness of VerbNet is further accentuated by available mappings~\cite{Loper:2007ws} to a number of other verb resources such as WordNet \cite{Fellbaum:1998wn}, FrameNet \cite{Baker:1998acl}, and PropBank \cite{Palmer:2005cl}.} The current version of VerbNet (v3.2) contains 8,537 distinct English verbs grouped into 273 VerbNet main classes. 

The inter-relatedness of syntactic behaviour and meaning of verbs is not limited to English~\cite{Levin:1993book}. The basic meaning components underlying verb classes are said to be \textit{cross-linguistically valid} \cite{Jackendoff:1992book,Merlo:2002acl}\footnote{For example, Levin \shortcite{Levin:1993book} notes that verbs in Warlpiri manifest analogous behavior to English with respect to the conative alternation. In another example, Polish verbs have the same patterns as \textsc{en} verbs in terms of the middle construction.} and therefore the classification has a strong cross-lingual dimension. A recent investigation of \newcite{Majewska:2017lre} show that it is possible to manually translate VerbNet classes and class members to different, typologically diverse languages with high accuracy.

The practical usefulness of VerbNet style classification both within and across languages has been limited by the fact that few languages boast resources similar to the English VerbNet. Some VerbNets have been developed completely manually from scratch, aiming to capture properties specific to the language in question, e.g., the resources for Spanish and Catalan \cite{Aparicio:2008lrec}, Czech \cite{Pala:2008raslan}, and Mandarin \cite{Liu:2008ll}. Other VerbNets were created semi-automatically, with the help of other lexical resources, e.g., for French \cite{Pradet:2014lrec} and Brazilian Portuguese \cite{Scarton:2012lrec}. These approaches involved substantial amounts of specialised linguistic and translation work. Finally, automatic methods have been developed, e.g., for French \cite{Sun:2010coling,Falk:2012acl} and Brazilian Portuguese \cite{Scarton:2014cicling}, with insufficient accuracy (as emphasised in Sect.~\ref{s:results}). Until now, work in this area has been limited to a small number of languages, due to the large requirements in terms of human input and/or the availability of mature NLP pipelines which exist only for a few resource-rich languages (e.g., English, German).

In this work, we propose a novel, fully automated approach for inducing VerbNets for multiple languages - one based on cross-lingual transfer. Unlike earlier approaches, our method does not require any parsed data or manual annotations for the target language. It encodes the cross-linguistic validity of Levin-style verb classifications into the vector-space specialisation framework (Sect.~\ref{ss:vss}) driven by linguistic constraints. A standard clustering algorithm is then run on top of the VerbNet-specialised representations using vector dimensions as features to learn verb clusters (Sect.~\ref{ss:clustering}). Our approach attains state-of-the-art verb classification performance across all six target languages.

\subsection{Vector Space Specialisation}
\label{ss:vss}
\paragraph{Specialisation Model}
Our departure point is a state-of-the-art specialisation model for fine-tuning vector spaces termed \textsc{Paragram} \cite{Wieting:2015tacl}.\footnote{The original \textsc{Paragram} model as well as other fine-tuning models from prior work inject pairwise linguistic constraints into existing vector spaces in order to improve their ability to capture semantic similarity/paraphrasability. In this work, we demonstrate that the same generic specialisation framework can be used to transform vector spaces to capture other types of lexical relations such as VerbNet relations.} The \textsc{Paragram} procedure injects similarity constraints between word pairs in order to make their vector space representations more similar; we term these the \textsc{Attract} constraints. Let $V=V_s \sqcup V_t$ be the vocabulary consisting of the source language and target language vocabularies $V_s$ and $V_t$, respectively. Let $C$ be the set of word pairs standing in desirable lexical relations; these include: \textbf{1)} verb pairs from the same VerbNet class (e.g.~(\emph{en\_transport, en\_transfer}) from verb class \textsc{send-11.1}); \iffalse or (\emph{en\_bark, en\_shout}) from the class \textsc{manner\_speaking-37.3}, \fi {and} \textbf{2)} the cross-lingual synonymy pairs (e.g.~(\textit{en\_{peace}, fi\_{rauha}})). Given the initial distributional space and collections of such \textsc{Attract} pairs $C$, the model gradually modifies the space to bring the designated word vectors closer together, working in mini-batches of size $k$. The method's cost function can be expressed as:

\vspace{-0.7em}
{\small
\begin{align}
O(\mathcal{B}_C) = O_{C}(\mathcal{B}_C) + R(\mathcal{B}_C)
\end{align}}%
The first term of the method's cost function (i.e., $\mathcal{O}_{C}$) pulls the \textsc{Attract} examples $(x_l, x_r) \in C$ closer together (see Fig.~\ref{fig:illustration} for an illustration). $B_{C}$ refers to the current mini-batch of \textsc{Attract} constraints. This term is expressed as follows: 

\vspace{-0.5em}
{\small
\begin{align} 
O_{C}(\mathcal{B}_C) ~=  \sum_{ (x_l, x_r) \in \mathcal{B}_{C}} & \big[ ~ \tau \left( \delta_{att} +  \mathbf{x}_l \mathbf{t}_l - \mathbf{x}_l \mathbf{x}_r \right)  \notag \\
+& \tau \left( \delta_{att} +  \mathbf{x}_r \mathbf{t}_r - \mathbf{x}_l \mathbf{x}_r  \right) \big]
\end{align}}%
$\tau(x)=\max(0,x)$ is the standard rectified linear unit or the hinge loss function \cite{Joachims:2004icml,Nair:2010icml}. $\delta_{att}$ is the ``attract'' margin: it determines how much vectors of words from \textsc{Attract} constraints should be closer to each other than to their negative examples. The negative example $\mathbf{t}_i$ for each word $x_i$ in any \textsc{Attract} pair is always the vector closest to $\mathbf{x}_i$ taken from the pairs in the current mini-batch, distinct from the other word paired with ${x}_i$, and ${x}_i$ itself.\footnote{Effectively, this term forces word pairs from the in-batch \textsc{Attract} constraints to be closer to one another than to any other word in the current mini-batch.}

The second $R(\mathcal{B}_C)$ term is the regularisation which aims to retain the semantic information encoded in the initial distributional space as long as this information does not contradict the used \textsc{Attract} constraints. Let $\mathbf{x}_{i}^{init}$ refer to the initial distributional vector of the word $x_i$ and let $\mathcal{V}(\mathcal{B}_C)$ be the set of all word vectors present in the given mini-batch. If $\lambda_{reg}$ denotes the L2 regularisation constant, this term can be expressed as:

\vspace{-0.5em}
{\small
\begin{align}
R(\mathcal{B}_C) =  \sum\limits_{ \mathbf{x}_i \in \mathcal{V}(\mathcal{B}_C)}  \lambda_{reg} \left\| \mathbf{x}^{init}_{i} - \mathbf{x}_i \right\|_{2} 
\end{align}}%

\paragraph{Linguistic Constraints: Transferring VerbNet-Style Knowledge} The fine-tuning procedure effectively blends the knowledge from external resources (i.e.,~the input \textsc{Attract} set of constraints) with distributional information extracted directly from large corpora. We show how to propagate annotations from a knowledge source such as VerbNet from source to target by combining two types of constraints within the specialisation framework: \textbf{a)} cross-lingual (translation) links between languages, and \textbf{b)} available VerbNet annotations in a resource-rich language transformed into pairwise constraints. Cross-lingual constraints such as \textit{(pl\_wojna, it\_guerra)} are extracted from BabelNet \cite{Navigli:12}, a large-scale resource which groups words into cross-lingual \textsc{babel} synsets (and is currently available for 271 languages). The wide and steadily growing coverage of languages in BabelNet means that our proposed framework promises to support the transfer of VerbNet-style information to numerous target languages (with increasingly high accuracy).

\begin{table}[t]
\centering
\vspace{-0.0em}
\def\arraystretch{0.96}
{\scriptsize
\begin{tabularx}{0.48\textwidth}{lll}
\toprule
{\footnotesize \textsc{learn-14}} & {\footnotesize \textsc{break-45.1}} & {\footnotesize \textsc{accept-77}}\\
\midrule
{(learn, study)} & {(break, dissolve)} & {(accept, understand)} \\
{(learn, relearn)} & {(crash, crush)} & {(accept, reject)} \\
{(read, study)} & {(shatter, split)} & {(repent, rue)} \\
{(cram, relearn)} & {(break, rip)} & {(reject, discourage)} \\
{(read, assimilate)} & {(crack, smash)} & {(encourage, discourage)} \\
{(learn, assimilate)} & {(shred, splinter)} & {(reject, discourage)} \\
{(read, relearn)} & {(snap, tear)} & {(disprefer, understand)} \\
\bottomrule
\end{tabularx}
}
\vspace{-0.5em}
\caption{Example pairwise \textsc{Attract} constraints extracted from three VerbNet classes in English.}
\label{tab:cons}
\vspace{-2.2mm}
\end{table}

To establish that the proposed transfer approach is in fact independent of the chosen cross-lingual information source, we also experiment with another cross-lingual dictionary: PanLex \cite{Kamholz:2014lrec}, which was used in prior work on cross-lingual word vector spaces \cite{Duong:2016emnlp,Adams:2017eacl}. This dictionary currently covers around 1,300 language varieties with over 12 million expressions, thus offering support also for low-resource transfer settings.\footnote{Similar to BabelNet, the translations in PanLex were derived from various sources such as glossaries, dictionaries, and automatic inference from other languages. This results in a high-coverage lexicon containing a certain amount of noise.}

VerbNet constraints are extracted from the English VerbNet class structure in a straightforward manner. For each class ${VN}_i$ from the 273 VerbNet classes, we simply take the set of all $n_i$ verbs $CL_i = \{v_{1,i}, v_{2,i}, \ldots, v_{n_i,i}\}$ associated with that class, including its subclasses, and generate all unique pairs $(v_k, v_l)$ so that $v_k,v_l \in CL_i$ and $v_k \neq v_l$. Example VerbNet pairwise constraints are shown in Tab.~\ref{tab:cons}. Note that VerbNet classes in practice contain verb instances standing in a variety of lexical relations, including synonyms, antonyms, troponyms, hypernyms, and the class membership is determined on the basis of connections between the syntactic patterns and the underlying semantic relations \cite{Kipper:2006lrec,Kipper:2008lre}.%\footnote{See also the VerbNet guidelines: \\ \scriptsize{http://verbs.colorado.edu/verb-index/VerbNet\_Guidelines.pdf}}

\subsection{Clustering Algorithm}
\label{ss:clustering}
Given the initial distributional or specialised collection of target language vectors $\mathbf{V_t}$, we apply an off-the-shelf clustering algorithm on top of these vectors in order to group verbs into classes. Following prior work \cite{Brew:2002emnlp,Sun:2009emnlp,Sun:2010coling}, we employ the MNCut spectral clustering algorithm \cite{Meila:2001aistats}, which has wide applicability in similar NLP tasks which  involve high-dimensional feature spaces \cite[i.a.]{Chen:2006acl,Luxburg:2007,Scarton:2014cicling}. Again, following prior work \cite{Sun:2010coling,Sun:2013acl}, we estimate the number of clusters $K_{Clust}$ using the self-tuning method of \newcite{Zelnik:2004nips}. This algorithm finds the optimal number by minimising a cost function based on the eigenvector structure of the word similarity matrix. We refer the reader to the relevant literature for further details. %regarding the clustering algorithm.

\section{Experimental Setup}
\label{s:exp}
\begin{table*}
\centering
\vspace{-0.0em}
\def\arraystretch{0.93}
{\footnotesize
\begin{tabularx}{\linewidth}{l llllll}
\toprule
{} & {\bf French:FR} & {\bf Portuguese:PT} & {\bf Italian:IT} & {\bf Polish:PL} & {\bf Croatian:HR} & {\bf Finnish:FI}\\
%{\bf Vectors} &  \multicolumn{2}{l}{} &  \multicolumn{2}{l}{} \\
\cmidrule(lr){2-7}
{\bf Training} & {} & {} & {} & {} & {} \\
{Corpus} & {frWaC} & {brWaC} & {itWaC} & {Araneum} & {hrWaC} & {fiWaC}\\
{Corpus size (in tokens)} & {1.6B} & {2.7B} & {2B} & {1.2B} & {1.4B} & {1.7B} \\
{Vocabulary size} & {242,611} & {257,310} & {178,965} & {373,882} & {396,103} & {448,863}\\
\midrule
{\bf Constraints} & {} & {} & {} & {} & {} \\
{\# monolingual VerbNet-\textsc{en}} & {220,052} & {220,052} & {220,052} & {220,052} & {220,052} & {220,052}\\
{\# mono \textsc{target} (\textsc{Mono-Syn})} & {428,329} & {292,937} & {362,452} & {423,711} & {209,626} & {377,548}\\
{\# cross-ling \textsc{en}-\textsc{target} (BNet)} & {310,410} & {245,354} & {258,102} & {219,614} & {160,963} & {284,167}\\
{\# cross-ling \textsc{en}-\textsc{target} (PLex)} & {225,819} & {187,386} & {216,574} & {154,159} & {201,329} & {257,106}\\
\midrule
{\bf Test} & {} & {} & {} & {} & {} \\
{\# Verbs (\# Classes)} & {169 (16)} & {660 (17)} & {177 (17)} & {258 (17)} & {277 (17)}  & {201 (17)}\\
{Coverage of test instances} & {94.1\%} & {95.5\%} & {96.6\%} & {93.4\%} & {98.2\%} & {84.6\%}\\
\bottomrule
\end{tabularx}
}
\vspace{-0.5em}
\caption{Statistics of the experimental setup for each target language: training/test data and constraints. \textit{Coverage} refers to the percentage of test verbs represented in the target language vocabularies.}
\vspace{-2.2mm}
\label{tab:stats}
\end{table*}

\paragraph{Languages}
We experiment with six target languages: French (\textsc{fr}), Brazilian Portuguese (\textsc{pt}), Italian (\textsc{it}), Polish (\textsc{pl}), Croatian (\textsc{hr}), and Finnish (\textsc{fi}). All statistics regarding the source and size of training and test data, and linguistic constraints for each target language are summarised in Tab.~\ref{tab:stats}. 

Automatic approaches to verb class induction have been tried out in prior work for \textsc{fr} and \textsc{pt}. To the best of our knowledge, our cross-lingual study is the first aiming to generalise an automatic induction method to more languages using an underlying methodology which is language-pair independent.

\paragraph{Initial Vector Space: Training Data and Setup} 
All target language vectors were trained on large monolingual running text using the same setup:  300-dimensional word vectors, the frequency cut-off set to 100, bag-of-words (BOW) contexts, and the window size of 2 \cite{Levy:2014acl,Schwartz:2016naacl}. All tokens were lowercased, and all numbers were converted to a placeholder symbol $<$NUM$>$.\footnote{Other SGNS parameters were also set to standard values \cite{Baroni:2014acl,Vulic:2016acluniversal}: $15$ epochs, $15$ negative samples, global learning rate: $.025$, subsampling rate: $1e-4$. Similar trends in results persist with $d=100,500$.}  \textsc{fr} and \textsc{it} word vectors were trained on the standard frWaC and itWaC corpora \cite{Baroni:2009lre}, and vectors for other target languages were trained on the corpora of similar style and size: \textsc{hr} vectors were trained on the hrWaC corpus \cite{Ljubesic:2014wac}, \textsc{pt} vectors on ptWaC \cite{Wagner:2016propor}, \textsc{fi} vectors on fiWaC \cite{Ljubesic:2016fiwac}, and \textsc{pl} vectors on the Araneum Polonicum Maius Web corpus \cite{Benko:2014tsd}. %The corpora and vocabulary sizes for each target language are provided in Tab.~\ref{tab:stats}. 
Note that we do not utilise any VerbNet-specific knowledge in the target language to induce and further specialise these word vectors.

Source \textsc{en} vectors were taken directly from the work of \newcite{Levy:2014acl}: they are trained with SGNS on the cleaned and tokenised Polyglot Wikipedia \cite{AlRfou:2013conll} containing $\sim$75M sentences, $\sim$1.7B word tokens and a vocabulary of $\sim$180k words after lowercasing and frequency cut-off. To measure the importance of the starting source language space as well as to test if syntactic knowledge on the source side may be propagated to the target space, we test two variant \textsc{en} vector spaces: SGNS with (a) BOW contexts and the window size 2 (SGNS-BOW2); and (b) dependency-based contexts (SGNS-DEPS) \cite{Pado:2007cl,Levy:2014acl}.%emphasising functional similarity \cite{Turney:2012jair}.

\paragraph{Linguistic Constraints}
We experiment with the following constraint types: (a) monolingual synonymy constraints in each target language extracted from BabelNet (\textit{Mono-Syn}); (b) cross-lingual \textsc{en}-\textsc{target} constraints from BabelNet; (c) cross-lingual \textsc{en}-\textsc{target} constraints plus \textsc{en} VerbNet constraints (see Sect.~\ref{ss:vss} and Fig.~\ref{fig:illustration}). Unless stated otherwise, we use BabelNet as the default source of cross-lingual constraints for (b) and (c).%The statistics regarding the numbers of available BabelNet and VerbNet constraints are summarised in Tab.~\ref{tab:stats}.  % constraints per target language

\paragraph{Vector Space Specialisation} 
The \textsc{Paragram} model's parameters are adopted directly from prior work \cite{Wieting:2015tacl} without any additional fine-tuning: $\delta_{att}=0.6$, $\lambda_{reg}=10^{-9}, k=50$. We train for 5 epochs without early stopping using AdaGrad \cite{Duchi:11}. \textsc{Paragram} is in fact a special case of the more general \textsc{Attract-Repel} specialisation framework \cite{Mrksic:2017tacl}: we use this more recent and more efficient TensorFlow implementation of the model in all experiments.\footnote{https://github.com/nmrksic/attract-repel}

\paragraph{Test Data} 
The development of an automatic verb classification approach requires an initial gold standard \cite{Sun:2010coling}: these have been developed for \textsc{fr} \cite{Sun:2010coling}, \textsc{pt} \cite{Scarton:2014cicling}, \textsc{it}, \textsc{pl}, \textsc{hr}, and \textsc{fi} \cite{Majewska:2017lre}. They were created using the methodology of \newcite{Sun:2010coling}, based on the \textsc{en} gold standard of \newcite{Sun:2008cicling} which contains 17 fine-grained Levin classes with 12 member verbs each. For instance, the class \textsc{put-9.1} in French contains verbs such as \textit{accrocher, d\'{e}poser, mettre, r\'{e}partir,  r\'{e}int\'{e}grer}, etc. %The test data is provided as supplemental material.
 %The statistics of the test data are provided in Tab.~\ref{tab:stats}. 

\paragraph{Evaluation Measures}
We use standard evaluation measures from prior work on verb clustering \cite[i.a.]{Diarmuid:2008coling,Sun:2009emnlp,Sun:2010coling,Falk:2012acl}. The mean \textit{precision} of induced verb clusters labelled \textit{modified purity} (\textsc{mPur}) is computed as:

\vspace{-1.1em}
{\small
\begin{align}
\text{\textsc{mPur}} = \frac{\sum_{C \in \mathbf{Clust}, n_{prev(C)}>1} n_{prev(C)}}{\#\text{test\_verbs}} 
\end{align}}%
Here, each cluster $C$ from the set of all $K_{Clust}$ induced clusters $\mathbf{Clust}$ is associated with its prevalent class/cluster from the gold standard, and the number of verbs in an induced cluster $C$ taking this prevalent class is labelled $n_{prev(C)}$. All other verbs not taking the prevalent class are considered errors.\footnote{Clusters with $n_{prev(C)}=1$ are discarded from the count to avoid an undesired bias towards singleton clusters \cite{Sun:2009emnlp,Sun:2010coling}.} \textit{\#{test\_verbs}} denotes the total number of test verb instances. The second measure targeting \textit{recall} is \textit{weighted class accuracy} (\textsc{wAcc}), computed as:
\begin{figure*}[t]
    \centering
    \begin{subfigure}[t]{0.495\linewidth}
        \centering
        \includegraphics[width=1.00\linewidth]{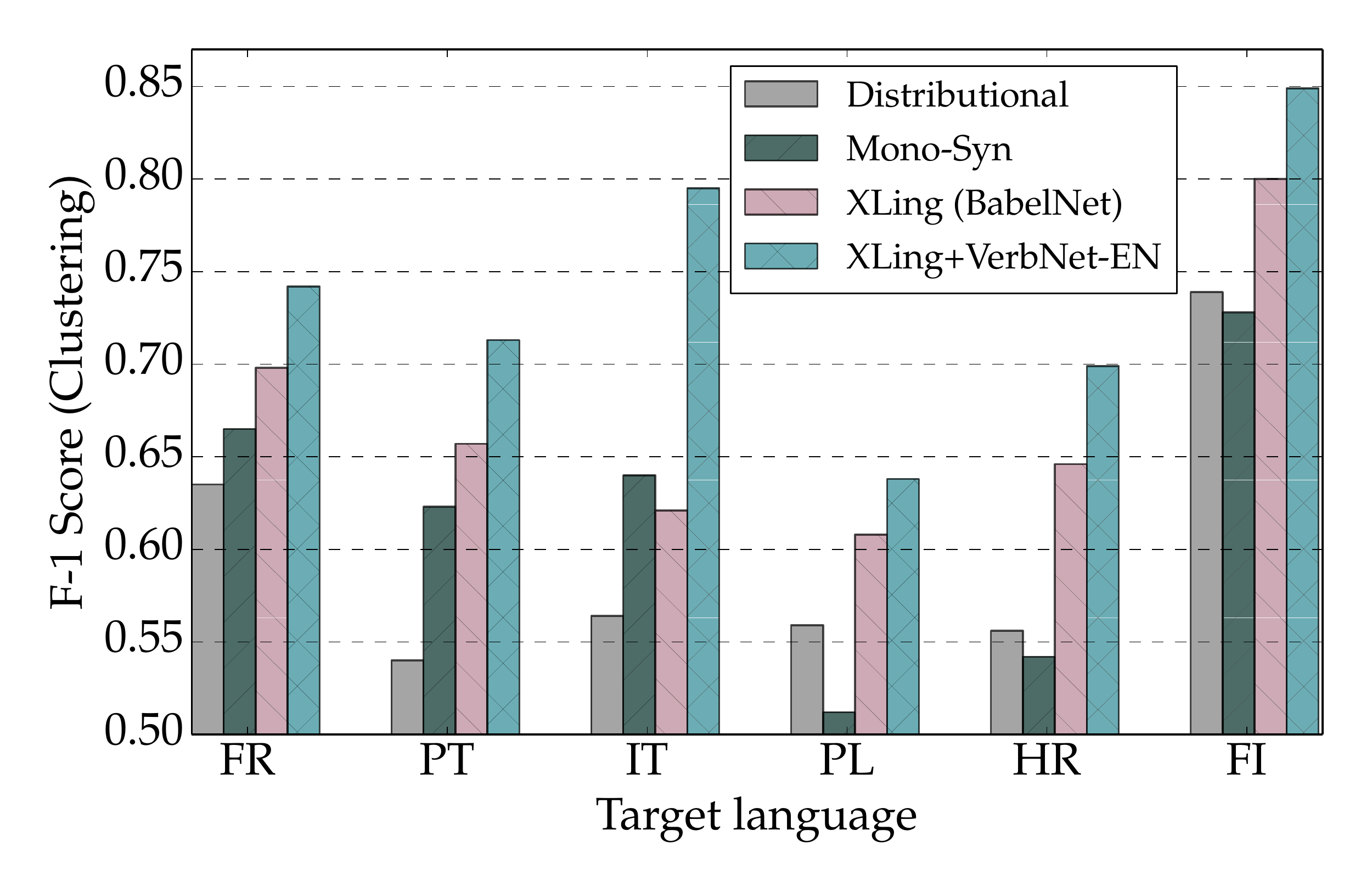}
        \caption{\textbf{EN Vectors: SGNS-BOW2}}
        \label{fig:bow}
    \end{subfigure}
    \begin{subfigure}[t]{0.495\textwidth}
        \centering
        \includegraphics[width=1.00\linewidth]{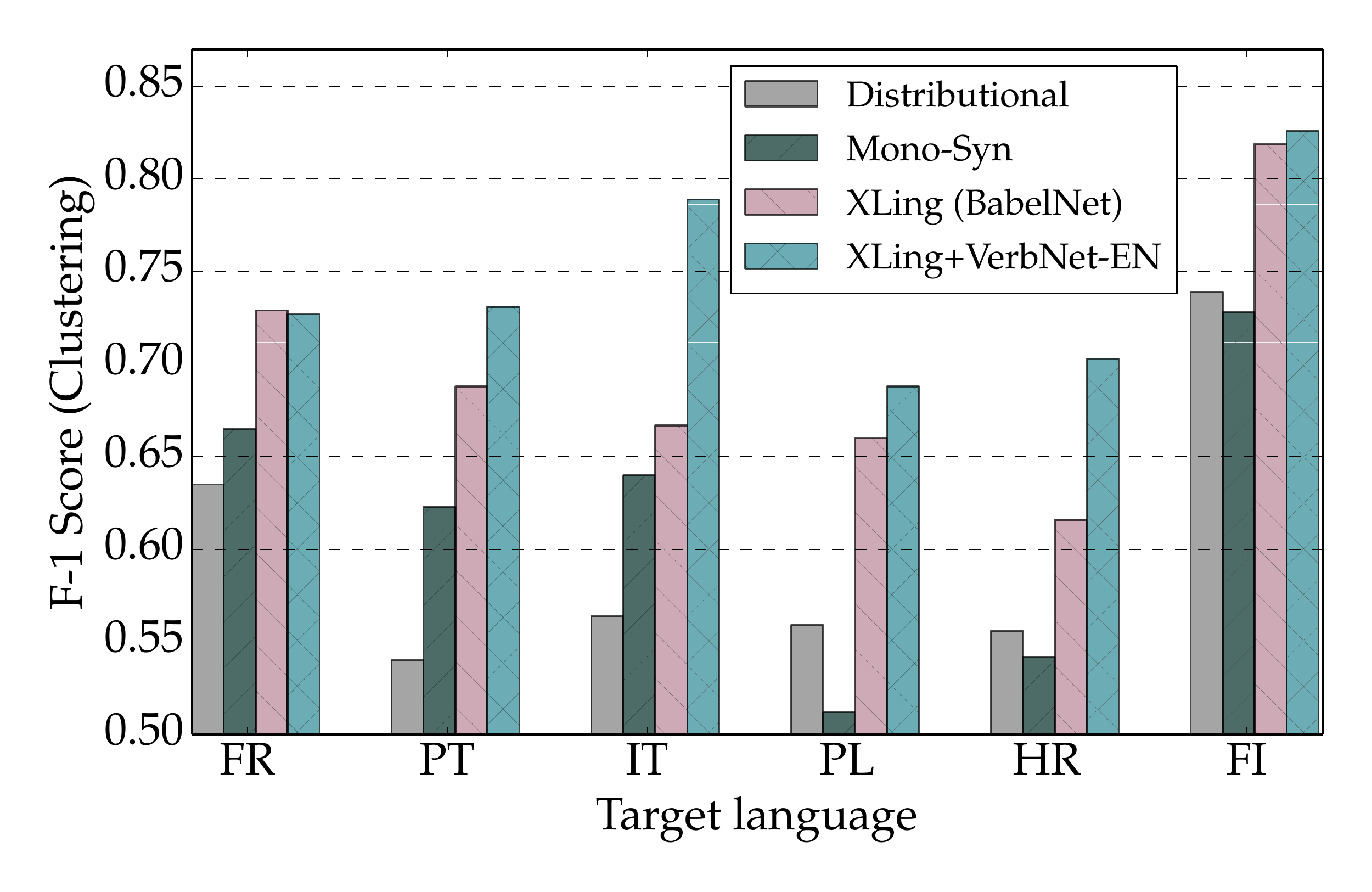}
        \caption{\textbf{EN Vectors: SGNS-DEPS}}
        \label{fig:deps}
    \end{subfigure}
    \vspace{-0.5em}
    %\caption{\textit{F-1} scores in six target languages using the fine-tuning post-processing \textsc{Attract} procedure from Sect.~\ref{ss:vss} and different sets of constraints: \textit{Distributional} refers to the initial vector space in each language without any further fine-tuning; \textit{Mono-Syn} refers to the vector space tuned using monolingual synonymy constraints from BabelNet; \textit{XLing} denotes the use of cross-lingual \textsc{en}-\textsc{target} constraints from BabelNet (where \textsc{target} refers to any of the six target languages); \textit{XLing+VerbNet-EN} is a fine-tuned vector space using both sets of constraints (see Tab.~\ref{tab:stats}): cross-lingual \textsc{en}-\textsc{target} constraints plus \textsc{en} VerbNet constraints, effectively imposing the VerbNet-based neighbourhoods in the vector subspace of the target language. Results are provided with (a) SGNS-BOW2 and (b) SGNS-DEPS source vector space in English for the \textit{XLing} and \textit{XLing+VerbNet} variants, see Sect.~\ref{s:exp}.}
\caption{\textit{F-1} scores in six target languages using the post-processing specialisation procedure from Sect.~\ref{ss:vss} and different sets of constraints: \textit{Distributional} refers to the initial vector space in each target language; \textit{Mono-Syn} is the vector space tuned using monolingual synonymy constraints from BabelNet; \textit{XLing} uses cross-lingual \textsc{en}-\textsc{target} constraints from BabelNet (\textsc{target} refers to any of the six target languages); \textit{XLing+VerbNet-EN} is a fine-tuned vector space which uses both cross-lingual \textsc{en}-\textsc{target} constraints plus \textsc{en} VerbNet constraints. Results are provided with (a) SGNS-BOW2 and (b) SGNS-DEPS source vector space in English for the \textit{XLing} and \textit{XLing+VerbNet} variants, see Sect.~\ref{s:exp}.}
\vspace{-2mm}
\label{fig:main}
\end{figure*}

\vspace{-1.1em}
{\small
\begin{align}
\text{\textsc{wAcc}} = \frac{\sum_{C \in \mathbf{Gold}} n_{dom(C)}}{\#\text{test\_verbs}} 
\end{align}}%
For each cluster $C$ from the set of gold standard clusters $\mathbf{Gold}$, we have to find the dominant cluster from the set of induced clusters: this cluster has the most verbs in common with the gold cluster $C$, and that number is $n_{dom(C)}$. As measures of precision and recall, \textsc{mPur} and \textsc{wAcc} may be combined into an F-1 score, computed as the balanced harmonic mean, which we report in this work.\footnote{We have also experimented with V-measure \cite{Rosenberg:2007emnlp}, another standard evaluation measure which balances between homogeneity (precision) and completeness (recall); we do not report these scores for brevity as similar trends in results are observed.} 

\section{Results and Discussion}
\label{s:results}
\noindent \textbf{Cross-Lingual Transfer Model~} F-1 verb classification scores for the six target languages with different sets of constraints are summarised in Fig.~\ref{fig:main}. We can draw several interesting conclusions. First, the strongest results on average are obtained with the model which transfers the VerbNet knowledge from English (as a resource-rich language) to the resource-lean target language (providing an answer to question Q3, Sect.~\ref{s:intro}). These improvements are visible across all target languages, empirically demonstrating the cross-lingual nature of VerbNet-style classifications. Second, using cross-lingual constraints alone (\textit{XLing}) yields strong gains over initial distributional spaces (answering Q1 and Q2). Fig.~\ref{fig:main} also shows that cross-lingual similarity constraints are more beneficial than the monolingual ones, despite a larger total number of the monolingual constraints in each language (see Tab.~\ref{tab:stats}). This suggests that such cross-lingual similarity links are strong implicit indicators of class membership. Namely, target language words which map to the same source language word are likely to be synonyms and consequently end up in the same verb class in the target language. However, the cross-lingual links are even more useful as means for transferring the VerbNet knowledge, as evidenced by additional gains with \textit{XLing+VerbNet-EN}.

% IV (removed): hypothesised by \cite{Levin:1993book,Majewska:2016msc}

The absolute classification scores are the lowest for the two Slavic languages: \textsc{pl} and \textsc{hr}. This may be partially explained by the lowest number of cross-lingual constraints for the two languages covering only a subset of their entire vocabularies (see Tab.~\ref{tab:stats} and compare the total number of constraints for \textsc{hr} and \textsc{pl} to the numbers for e.g. \textsc{fi} or \textsc{fr}). Another reason for weaker performance of these two languages could be their rich morphology, which induces data sparsity both in the initial vector space estimation and in the coverage of constraints.

\paragraph{State-of-the-Art}
A direct comparison of previous state-of-the-art classification scores available for \textsc{fr} \cite{Sun:2010coling} and \textsc{pt} \cite{Scarton:2014cicling} on the same test data exemplifies the extent of improvement achieved by our transfer model. F-1 scores improve from 0.55 to 0.75 for \textsc{fr} and from 0.43 to 0.73 for \textsc{pt}. \newcite{Scarton:2014cicling} explain the low performance by ``the lower quality NLP tools''. This issue is largely mitigated by our VerbNet transfer model, which exploits the assumption of cross-linguistic class consistency directly through a specialised vector space, and also avoids any reliance on target-language-specific NLP tools. 

\paragraph{Starting Source Vector Space}
Fig.~\ref{fig:bow} and Fig.~\ref{fig:deps} enable a brief analysis of the influence of the starting \textsc{en} vector space on the results for each target language. We observe small but consistent gains with SGNS-DEPS, which utilises syntactic information stemming from a dependency parser on the source side, over SGNS-BOW for the \textit{XLing} variant. The improvements are +2.1 points on average, visible for 5 out of 6 target languages. We again see an increase in performance with the \textit{XLing+VerbNet} model, but we do not observe any major difference between the two starting source spaces now: average slight score difference of 0.3 is in favour of SGNS-BOW2, which outperforms SGNS-DEPS for 3 out of 6 target languages. This finding indicates that VerbNet-based linguistic constraints are more important for the final classification performance, and mitigate the artefacts of the starting distributional source space.

\begin{table}
\centering
\vspace{-0.0em}
\def\arraystretch{0.91}
{\footnotesize
\begin{tabularx}{\linewidth}{l ll ll}
\toprule
{} & \multicolumn{2}{l}{\bf French} & \multicolumn{2}{l}{\bf Italian} \\
\midrule
{\scriptsize \bf Vector space} & {\footnotesize \em XLing} & {\footnotesize \em XLing+VN} & {\footnotesize \em XLing} & {\footnotesize \em XLing+VN} \\
\cmidrule(lr){2-3} \cmidrule(lr){4-5}
{\textsc{en}+\textsc{fr}} & {0.698} & {0.742} & {--} & {--} \\
{\textsc{en}+\textsc{it}} & {--} & {--} & {0.621} & {\bf 0.795} \\
{\textsc{en}+\textsc{fr}+\textsc{it}} & {\bf 0.728} & {0.745} & {0.650} & {0.768} \\
{\textsc{en}+\textsc{fr}+\textsc{pl}} & {0.697} & {0.717} & {--} & {--} \\
{\textsc{en}+\textsc{it}+\textsc{pl}} & {--} & {--} & {0.658} & {0.765} \\
{\scriptsize \textsc{en}+\textsc{fi}+\textsc{fr}+\textsc{it}} & {0.719} & {\bf 0.751} & {0.662} & {0.777} \\
{\scriptsize \textsc{en}+\textsc{fr}+\textsc{it}+\textsc{pt}} & {0.710} & {0.718} & {\bf 0.688} & {0.760} \\

\bottomrule
\end{tabularx}
}
\vspace{-0.3em}
\caption{The effect of multilingual vector space specialisation. Results are reported for \textsc{fr} and \textsc{it} using: a) cross-lingual constraints only (\textit{XLing}); and b) the VerbNet transfer model (\textit{XLing+VN}).}
\vspace{-1mm}
\label{tab:multi}
\end{table}

\paragraph{Bilingual vs. Multilingual} The transfer model can operate with more than two languages, effectively inducing a multilingual vector space. We analyse such multilingual training based on the results on \textsc{fr} and \textsc{it} (Tab.~\ref{tab:multi}). On average, the results with \textit{XLing} improve with more languages (see also the results for \textsc{en} in Tab.~\ref{tab:vcsim}), as the model relies on more constraints for the vector space specialisation. Yet additional languages do not lead to clear improvements with \textit{XLing+VerbNet}: we hypothesise that the specialisation procedure becomes dominated by cross-lingual constraints which may diminish the importance of VerbNet-based \textsc{en} constraints. The language configuration in the multilingual vector space also makes a difference: e.g., combining \textsc{pl} with the Romance languages degrades the performance, while \textsc{fi} surprisingly boosts it slightly. For brevity, we only report the results for \textsc{fr} and \textsc{it}. Similar trends are observed when making the transition from bilingual to multilingual vector spaces for other target languages.

%\paragraph{Cross-Lingual Constraints}

%\paragraph{Full Transfer Model}

\begin{table}
\centering
\vspace{-0.0em}
\def\arraystretch{0.9}
{\footnotesize
\begin{tabularx}{\linewidth}{l lll}
\toprule
{} & {\textit{VC}:XLing} & {\textit{Sim}:XLing} & {\textit{Sim}:XLing+VN} \\
\midrule
{\textsc{en}-Dist} & {0.484} & {0.275} & {--} \\
\midrule
{+\textsc{fr}} & {0.608} & {0.556} & {0.481} \\
{+\textsc{pt}} & {0.633} & {0.537} & {0.466} \\
{+\textsc{it}} & {0.602} & {0.524} & {0.476} \\
{+\textsc{pl}} & {0.597} & {0.469} & {0.431} \\
{+\textsc{hr}} & {0.582} & {0.497} & {0.446} \\
{+\textsc{fi}} & {0.662} & {0.598} & {0.491} \\
\midrule
{+\textsc{fr}+\textsc{it}} & {0.633} & {0.571} & {0.526} \\
%{+\textsc{fr}+\textsc{pl}} & {0.619} & {0.565} & {0.513} \\
%{+\textsc{it}+\textsc{pl}} & {0.671} & {0.540} & {} \\
{+\textsc{fi}+\textsc{fr}+\textsc{it}} & {0.641} & {0.635} & {0.558} \\
{+\textsc{fr}+\textsc{it}+\textsc{pt}} & {0.674} & {0.596} & {0.515} \\
\midrule
{\textsc{en}-VN} & {\bf 0.956} & {--} & {0.358} \\

\bottomrule
\end{tabularx}
}
\vspace{-0.3em}
\caption{Comparison of verb classification (\textit{VC}) and verb semantic similarity (\textit{Sim}) for English. \textit{VC} is measured on the \textsc{en} test set of \newcite{Sun:2008cicling}. \textit{Sim} is measured on SimVerb-3500 \cite{Gerz:2016emnlp}. The scores are Spearman's $\rho$ correlation scores. \textsc{en}-Dist is the initial distributional English vector space: SGNS-BOW2; \textsc{en}-VN is the same space transformed using monolingual \textsc{en} VerbNet constraints only, an upper bound for the specialisation-based approach in \textsc{en}.} %\textit{XLing} and \textit{XLing+VN} again denote two specialisation model variants.}
\vspace{-1mm}
\label{tab:vcsim}
\end{table}

%\paragraph{Other Source Vector Space}

%\paragraph{Further Discussion} 

\paragraph{Clustering Algorithm} 
Since vector space specialisation is detached from the application of the clustering algorithm, our framework allows straightforward experimentation with other algorithms. Following prior work \cite{Brew:2002emnlp,Sun:2010coling}, we also test K-means clustering. Results for the six languages using the \textsc{en }SGNS-BOW2 source space and \textit{Xling+VerbNet-EN} are on average 3.8 points lower than the ones reported in Fig.~\ref{fig:bow}. K-Means is outperformed for each target language, confirming the superiority of spectral clustering established in prior work, e.g., \cite{Scarton:2014cicling}. On the other hand, we find results with another clustering algorithm, hierarchical agglomerative clustering with Ward's linkage \cite{Ward:1963}, on par with spectral clustering (1.4 points on average in favour of spectral, which is better on 4 out of 6 languages). We believe that further gains in verb class induction could be achieved by additional fine-tuning of the clustering algorithm. %applied, but leave this for future work.

\begin{figure}[t]
\centering
\includegraphics[width=0.97\linewidth]{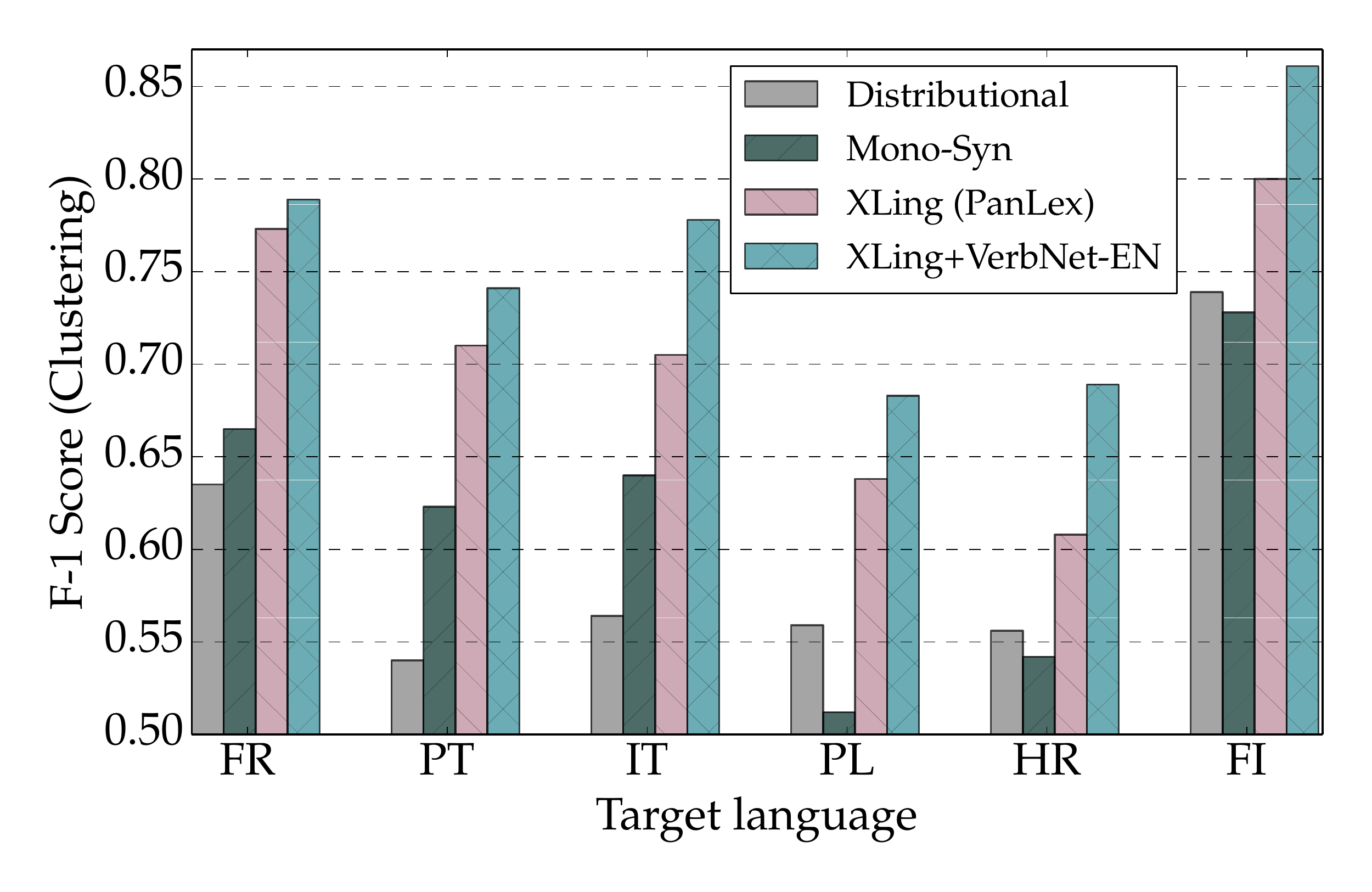}
\vspace{-3.5mm}
\caption{\textit{F-1} scores when PanLex is used as the source of cross-lingual \textsc{Attract} constraints (instead of BabelNet). EN Vectors: SGNS-BOW2. }
\vspace{-1mm}
\label{fig:panlex}
\end{figure}

\paragraph{Other Cross-Lingual Sources}
Replacing BabelNet with PanLex as the alternative source of cross-lingual information again leads to large gains with the cross-lingual transfer model, as is evident from Fig.~\ref{fig:panlex}. This suggests that the proposed approach does not depend on a particular source of information - it can be used with any general-purpose bilingual dictionary. We mark slight improvements for 3/6 target languages when comparing the results with the ones from Fig.~\ref{fig:bow}. The new state-of-the-art F-1 scores are 0.79 for \textsc{fr} and 0.74 for \textsc{pt}. 

\paragraph{Verb Classification vs. Semantic Similarity} 
An interesting question originating from prior work on verb representation learning, e.g., \cite{Baker:2014emnlp} touches upon the correlation between verb classification and semantic similarity. Due to the availability of VerbNet constraints and a recent similarity evaluation set (SimVerb-3500; it contains human similarity ratings for 3,500 verb pairs) \cite{Gerz:2016emnlp}, we perform the analysis on English: the results are summarised in Tab.~\ref{tab:vcsim}. They clearly indicate that cross-lingual synonymy constraints are useful for both relationship types (compare the scores with \textit{XLing}), with strong gains over the non-specialised distributional space. However, the inclusion of VerbNet information, while boosting classification scores for target languages and (trivially) for \textsc{en}, deteriorates \textsc{en} similarity scores across the board (compare \textit{XLing+VN} against \textit{XLing} in Tab.~\ref{tab:vcsim}). This suggests that the VerbNet-style class membership is definitely not equivalent to pure semantic similarity captured by SimVerb. 

\subsection{Further Discussion and Future Work}
This work has proven the potential of transferring lexical resources from resource-rich to resource-poor languages using general-purpose cross-lingual dictionaries and bilingual vector spaces as means of transfer within a semantic specialisation framework. However, we believe that the proposed basic framework may be upgraded and extended across several research paths in future work.

First, in the current work we have operated with standard single-sense/single-prototype representations, thus effectively disregarding the problem of verb polysemy. While several polysemy-aware verb classification models for English were developed recently \cite{Kawahara:2014acl,Peterson:2016sem}, the current lack of polysemy-aware evaluation sets in other languages impedes this line of research. Evaluation issues aside, one idea for future work is to use the \textsc{Attract-Repel} specialisation framework for sense-aware cross-lingual transfer relying on recently developed multi-sense/prototype word representations \cite[inter alia]{Neelakantan:2014emnlp,Pilehvar:2016emnlp}. 

Another challenge is to apply the idea from this work to enable cross-lingual transfer of other structured lexical resources available in English such as FrameNet \cite{Baker:1998acl}, PropBank \cite{Palmer:2005cl}, and VerbKB \cite{Wijaya:2016naacl}. Other potential research avenues include porting the approach to other typologically diverse languages and truly low-resource settings (e.g., with only limited amounts of parallel data), as well as experiments with other distributional spaces, e.g. \cite{Melamud:2016conll}. Further refinements of the specialisation and clustering algorithms may also result in improved verb class induction.

\section{Conclusion}
\label{s:conc}
We have presented a novel cross-lingual transfer model which enables the automatic induction of VerbNet-style verb classifications across multiple languages. The transfer is based on a word vector space specialisation framework, utilised to directly model the assumption of cross-linguistic validity of VerbNet-style classifications. Our results indicate strong improvements in verb classification accuracy across all six target languages explored. All automatically induced VerbNets are available at:\\ {\normalsize \texttt{github.com/cambridgeltl/verbnets}.

%\section{Related Work}
%\label{s:rw}
%\input{0x_rw}

\section*{Acknowledgments}
This work is supported by the ERC Consolidator Grant LEXICAL: Lexical Acquisition Across Languages (no 648909). The authors are grateful to the entire LEXICAL team, especially to Roi Reichart, and also to the three anonymous reviewers for their helpful and constructive suggestions.

\bibliography{emnlp2017_refs}
\bibliographystyle{emnlp_natbib}

\end{document}